\documentclass[letterpaper, 10 pt, conference]{IEEEtran}  

\IEEEoverridecommandlockouts                              

\usepackage{graphics} 
\usepackage{epsfig} 
\usepackage{mathptmx} 
\usepackage{times} 
\usepackage{amsmath} 
\usepackage{amssymb}  

\usepackage[T1]{fontenc}
\usepackage{amsfonts}
\usepackage{booktabs}
\usepackage{siunitx}
\usepackage{caption} 
\usepackage{booktabs}
\usepackage{graphicx}
\usepackage{algorithm, algpseudocode}
\usepackage{float}
\usepackage[percent]{overpic}
\usepackage{caption}
\DeclareGraphicsExtensions{.pdf,.png}

\usepackage[font=small,skip=0pt]{subcaption}

\usepackage{array, tabularx, makecell, booktabs}%

\usepackage{geometry}
\usepackage{comment}

\title{\LARGE \bf
Imagination-augmented Navigation Based on 2D Laser Sensor Observations
}

\author{Zhengcheng Shen$^{1}$, Linh K{\"a}stner$^{1}$\thanks{$^{1}$ Zhengcheng Shen , Linh K{\"a}stner, Magdalena Yordanova, and Jens Lambrecht are with the Chair Industry Grade Networks and Clouds, Faculty of Electrical Engineering, and Computer Science,				
		Berlin Institute of Technology, Berlin, Germany
		{\tt\small zhengcheng.shen@tu-berlin.de}}, Magdalena Yordanova$^{1}$, and Jens Lambrecht$^{1}$
}

\begin{document}

\maketitle
\thispagestyle{empty}
\pagestyle{empty}


\begin{abstract}
Autonomous navigation of mobile robots is an essential task for various industries. Sensor data is crucial to ensure safe and reliable navigation. However, sensor observations are often limited by different factors. Imagination can assist to enhance the view and aid navigation in dangerous or unknown situations where only limited sensor observation is available. In this paper, we propose an imagination-enhanced navigation based on 2D semantic laser scan data. The system contains an imagination module, which can predict the entire occupied area of the object. The imagination module is trained in a supervised manner using a collected training dataset from a 2D simulator. Four different imagination models are trained, and the imagination results are evaluated. Subsequently, the imagination results are integrated into the local and global cost map to benefit the navigation procedure. The approach is validated on three different test maps, with seven different paths for each map. The quality and numeric results showed that the agent with the imagination module could generate more reliable paths without passing beneath the object, with the cost of a longer path and slower velocity.

\end{abstract}


\section{Introduction}
\begin{figure}[]
    \centering
    \includegraphics[width=0.45\textwidth]{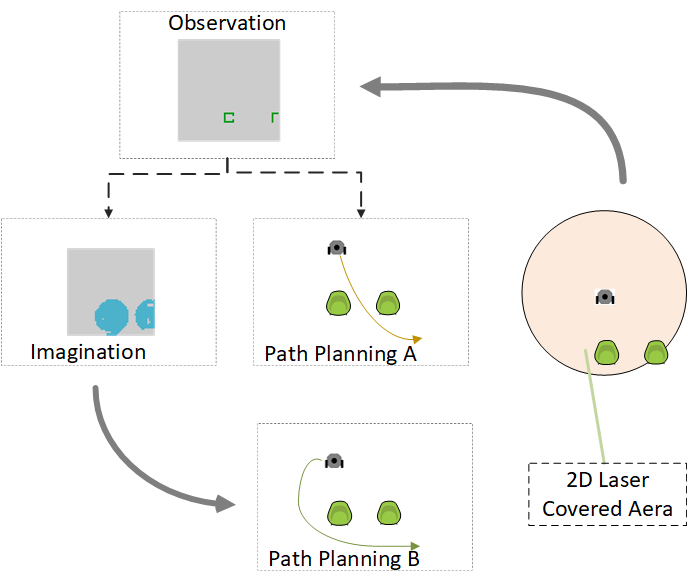}
    \caption{The proposed imagination module will alter the behavior of the path planner. The orange circle represents the laser scanner's observable area, while the robot only detects the legs of the chair. With the limited sensor information, the robot will decide to go through the chair, which may have a high risk to cause a collision or disturb persons. Using imagination, the robot will evaluate the path more carefully and choose a path with more adjustable space.}
    \label{intro}
\end{figure}

\noindent
In today's navigation systems, accurate and complete sensor information that can observe a wide area of the environment is crucial to ensure safe and reliable navigation \cite{robla2017working}, \cite{fragapane2020increasing} and a variety of different sensors are deployed to ensure safety robustness and redundancy. However, this can not always be guaranteed. Since the laser can not go through solid objects, there will be occluded areas \cite{DBLP:journals/corr/SongYZCSF16} which may not be seen by the robot. Moreover, the installation height of the sensor will also strongly limit the observation. For instance, for 2D laser sensors in a lower position, only the legs of the chair or table will be detected. Moreover, in hazardous environments, or when the connection is low, precious information could go missing. In these scenarios, imagination could substantially help to recount old memories to rebuild and/or augment the field of view and aid for navigation tasks \cite{DBLP:journals/corr/abs-2008-09285}.
Additionally, with the imagination enhanced observation, semantic information can be leveraged to achieve high-level path planning tasks \cite{DBLP:journals/corr/abs-2007-00643}, such as to avoid sensitive areas, e.g the area under tables or between chairs in specific situations like in Fig. \ref{intro}. The potential benefits of the imagination for navigation are further explored in the paper.
\noindent
In this paper, we propose an approach to train and deploy an imagination module to enhance navigational safety and efficiency based on 2D sensor data observations. We train four different neural network models to detect and imagine real-world objects from limited sensor observations. A dataset, which contains the observation and 4 paired ground truths for different models, is extracted from the arena-ros for the training of the imagination module. The input of the imagination unit will be a 2D local egocentric map. The resulting observation is directly merged into the occupancy map and provides the robot with an augmented view of its environment to improve planning. 

\noindent
The main contributions of this work are the following:
\begin{itemize}
    \item Proposal of an imagination module to semantically detect and imagine real-world objects based solely on 2D laser scan data. Various filters and preprocessing steps are introduced to convert an egocentric 2D laser scan into a semantic top-down map for training. Subsequently, four different network models are trained.

    \item Integration of the imagination module into the robot operating system (ROS). The imagination results will directly be merged into the occupancy grid maps used by global and local planners.
    
    \item Evaluation and validation of the imagination models on three different maps 

\end{itemize}

The paper is structured as follows. Sec. II begins with related works. Subsequently, the methodology is presented in Sec. III. Sec. IV presents the results and discussion. Finally, Sec. V will give a conclusion and outlook. We made the code publicly available under https://github.com/ignc-research/all-in-one-DRL-planner.

\section{Related Works}
\noindent
Autonomous navigation of mobile robots has been subject to a vast amount of research works \cite{chiang2019learning}, \cite{faust2018prm}, \cite{francis2020long}, \cite{long2017deep}. The majority of the work assumes complete sensor information, which is not always guaranteed in real-world settings. Thus, leveraging semantic information to additionally improve navigation has become an important aspect and has been explored by a variety of different research works in recent years. 
Everett et al. \cite{everett2019planning} proposed a navigation system that learns to navigate based on specific context learned by analyzing terrain context from semantic grid maps.
Narasimhan et al. \cite{narasimhan2020seeing} utilize semantic maps using amodal top-down maps and learn the architectural shape of rooms. The researchers demonstrate that the agent could learn to navigate to specific rooms in unknown homes without seeing the entire environment.
Works by Chen et al. \cite{chen2020soundspaces}, \cite{chen2021semantic}, \cite{chen2020learning} or Yu et al. \cite{yu2022sound} or Krantz et al. \cite{krantz2020beyond} have explored semantic navigation for high-level tasks such as finding specific objects, navigating and finding rooms based on visual and audio semantics.
Another work that focuses on high-level semantic navigation is proposed by Batra et al. \cite{batra2020objectnav}. The researchers present methods and benchmarks to assess and evaluate navigation based on high-level information of surrounding objects. They conclude and stress the importance of merging semantic information to contribute to intelligent navigation. 
\\
\noindent
Another aspect to be considered is that oftentimes, information is not perfectly given or the sensor does not work properly due to limited observations. Thus, semantic scene completion has also gained increasing interest in the research community. 
Song et al. \cite{song2018im2pano3d} proposed an approach to generate semantic 360 panoramas from partial observations of less than 50 percent. the authors leveraged a convolutional neural network to learn priors from a large-scale dataset.
Other researchers combined this semantic scene completion with navigation.
Liang et al. \cite{liang2021sscnav} proposed SSCNav to complete scenes with unknown obstacles to improve a DRL agent. the authors compute the agent's observation online by deploying a semantic scene completion module to enhance the agent's observations. The researchers demonstrated efficient and more reliable trajectories in unknown scenes.
Georgakis et al. \cite{georgakis2021learning} proposed an approach to actively learn to generate semantic maps of the environment outside the agent's field of view. The authors project the depth and RGB image into an egocentric top-down view to train a model that learns to predict semantics of unobserved areas in a supervised manner. The model could learn to predict semantically plausible scenes such as chairs around tables.
\noindent
Most of the aforementioned papers deal with photo-realistic 3D environments and visual information as they contain the most information. However, it is not always possible to have a visual sensor, and processing this data is computationally intensive. On that account, our work focuses on the usage of more lightweight sensor data such as 2d laser scan data. we utilize 2d laser scan data and enhance it with semantic information to improve navigation.

\begin{figure*}[!htp]
	\centering
	\includegraphics[width=0.96\textwidth]{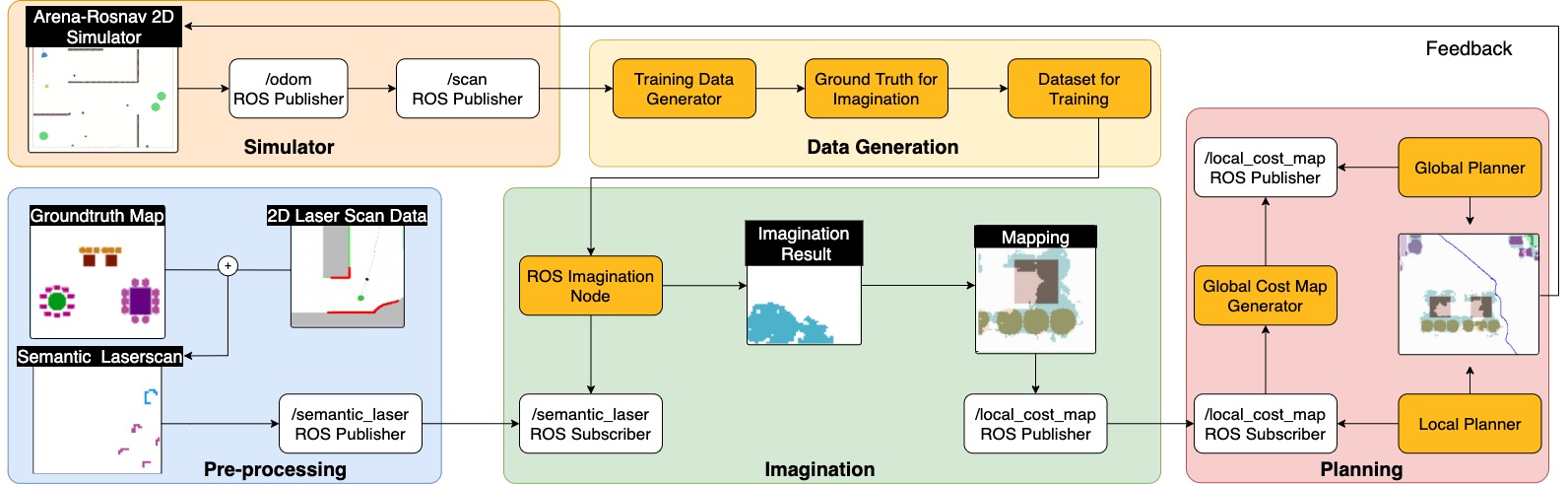}
	\caption{System design of our approach. The training data is acquired using a 2D simulator arena-rosnav \cite{kastner2021}. Preprocessing steps convert the egocentric 2D laserscan message into a semantic top-down map using the ground truth map. Subsequently, the semantic laser is given as input to the imagination node to detect and imagine the objects. Finally, the resulting occupancy of objects is directly merged into the local and global costmaps, which are consumed by the global and local planner. The robot operating system (ROS) is used as communication middleware.}
	\label{system}
\end{figure*}

\section{Methodology}
\noindent In this chapter, we present the methodology of our proposed framework. The main objective is to detect and imagine different object classes solely from 2d laser scan data and directly integrate this additional information for path planning. Our system design is illustrated in Fig. \ref{system}. 
We utilize a 2D simulation environment arena-rosnav, which simulates the robot and laser scan data. Using the simulator, training data is first generated and prepared for training. More specifically, the 2D egocentric laserscan information is transformed into a semantic top-down view map using the ground truth map. The network then learns to imagine the complete shape of the object from limited sensor observation and outputs an occupancy grid. This additional occupancy information is then merged with the existing global and local cost map of the ROS navigation stack, which is used for the global and local planner respectively. 
In the following, each module is described in more detail.

\subsection{Simulation Platform}
\noindent This work is utilizing arena-rosnav \cite{kastner2021} an efficient 2D simulation environment for navigation integrated into ROS. It is based on the open-source package Flatland, which allows simulation of dynamic and static 2D objects and implementation of ROS-based functionalities like sensor data, navigation models, etc. In this work, we spawn different chairs and tables that are classified using a specific ID. This ID is later used to semantically generate the semantic laser scan information. 
The middleware is ROS and respective information is gathered through topics and services. 
Since it's full integration into ROS, conventional planning methods can be used, which will be later on utilized to validate our approach. Furthermore, arena-rosnav also consists of tools to generate new worlds and scenarios, record data, and test navigation approaches. In this paper, a Turtlebot3 robot with a laser height of 0.2m is deployed. Several different maps consisting of static obstacles such as tables and chairs are spawned. Due to the low height of the laser scan sensor, the robot is only able to perceive the legs of objects like tables and chairs. Within this paper, our objective is to utilize the proposed imagination module to augment the limited observation space using imagination.

\subsection{Data Acquisition and Preprocessing}
\label{sec:data_acquisition}
\begin{figure}[!h]
	\centering
	\includegraphics[width=0.4\textwidth]{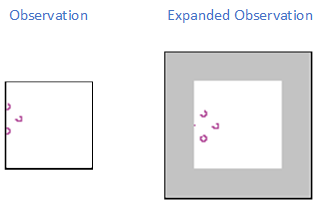}
	\caption{The normal observation vs extended observation}
	\label{fig:expanded_ob}
\end{figure}

\noindent The capability to get the semantic map from imagination is proved by \cite{DBLP:journals/corr/abs-2104-03638}. In this paper, we focus on the navigation influence created by the imagination. To integrate the system with arena-rosnav, the 2D laser scan data will be generated according to the following steps. 2D laser scan data is acquired using ROS topics. The simulated robot will provide a stream of 2d sensor data together with its position on the map. Within the ground truth map, all obstacles are assigned with a unique ID, which specifies their class. Using this information, each laser beam can be assigned an additional semantic id of the object it is perceiving. Thus, the egocentric laser scan observation of the robot can be semantically labeled to provide a top-down semantic map. Therefore, the ROS scan message type is extended by an ID specifying the respective class. This will then be used by the imagination unit. The procedure is illustrated in the pre-processing unit in Fig. \ref{system}.
To consider noise and improve the semantic object detection capability as presented in \cite{DBLP:journals/corr/abs-2104-03638}, we cut the laser sensor into a 60x60 top view image, which corresponds to a 3-meter by 3- meter area. Using imagination, the unseen area can be anticipated. For the imagination beyond the scope, we have to extend the 60x60 sensor observation into a 100x100 sensor observation, as illustrated in Fig. \ref{fig:expanded_ob}. We fill the ID value of unknown environments for the expanded area. For this expanded observation space, a different model will be trained.

\subsection{Imagination Unit}
\subsubsection{Unet}
\noindent The imagination unit consists of two different parts and is illustrated in Fig \ref{imagination-detail}. The first part is an Unet with 32 feature channels \cite{ronneberger2015u}. After the semantic laser scan data is received, it will be put into 4 down-sampling layers and generate an array of features $[x_1, x_2, x_3, x_4, x_5]$. The array of features will be given as input into the Unet decoder for imagination.
\begin{table}
\begin{center}
\caption{\label{tab:features}Features From Unet Encoder}
\begin{tabular}{ |c|c|c|c|c| } 
 \hline
 Features& Channels & Number 0f Features & Size \\ 
 $x_1$ & 32 & 32 & 60x60 \\ 
 $x_2$ & 32 & 64 & 30x30\\ 
 $x_3$ & 32 & 128 & 15x15\\ 
 $x_4$ & 32 & 256 & 7x7\\ 
 $x_5$ & 32 & 256 & 3x3\\ 
 \hline
\end{tabular}
\end{center}
\end{table}
\noindent The encoder is responsible for extracting patterns for imagination from the semantic laser scan data and forward the features into the Unet decoder. The decoder will take the features and give out the raw imagination of the local environment. 

\subsubsection{Post-process and Filters}
\begin{figure*}[!h]
	\centering
	\includegraphics[width=0.95\textwidth]{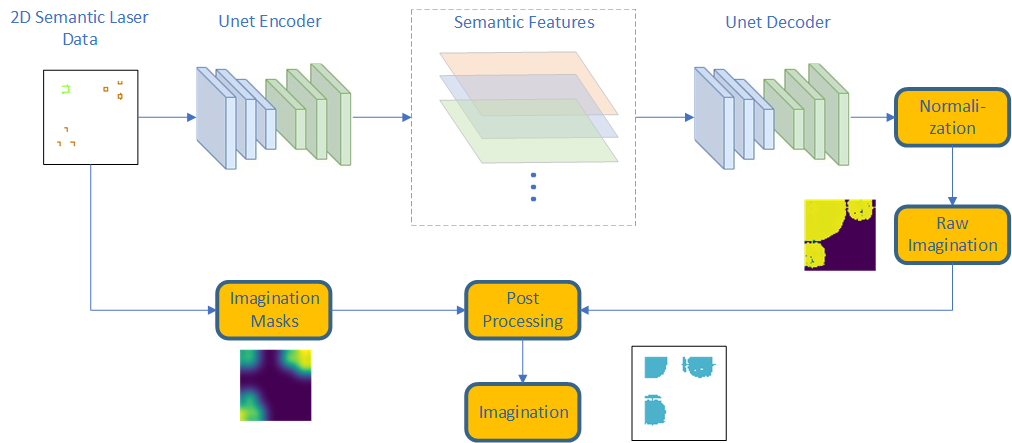}
	\caption{Detailed structure of the imagination unit. The imagination unit consists of an Unet encoder, an Unet decoder, the imagination mask generator, and the post-processing unit. The input data, 2D semantic laser data will first go through the Unet. The raw imagination output, combined with an input-based imagination mask will be then forwarded to the post-processing module for the final result.}
	\label{imagination-detail}
\end{figure*}

\noindent Another important part of the imagination Unit is the post-processing module depicted in Fig. \ref{imagination-detail}. Like humans, the imagination is usually accompanied by noise from relevant neuron activation. Sometimes it will suggest that there are objects even in free space. The noise from irrelevant neurons can be significantly reduced by deploying filters as shown in our previous works \cite{DBLP:journals/corr/abs-2104-03638}. The filter is created based on the sensor data and no ground truth data is involved. For each local laser scan map $M_{scan}$, the following procedure is executed:
\begin{equation}
    filter = conv(M_{scan}, kernel^{gauss})
\end{equation}
where $kernel^{gauss}$ is a 31x31 gaussian kernel where:
\begin{equation}
    kernel^{gauss}_{x,y} = \frac{1}{2\pi\sigma^2}e^{-\frac{x^2 + y^2}{2\sigma^2}} \\
\end{equation}

\noindent The result of our imagination module is a probability occupancy map, which is used to enhance the existing global and local occupancy maps of the navigation stack. A threshold $\theta$ will be set to control the intensity of the imagination, and all the points with a value lager than $\theta$ are labeled as occupied. The threshold in this paper is set to 0.2. The final result will be used to calculate the cost map used by the path planners.

\subsection{Training Procedure}
\begin{figure}[!h]
	\centering
	\includegraphics[width=0.4\textwidth]{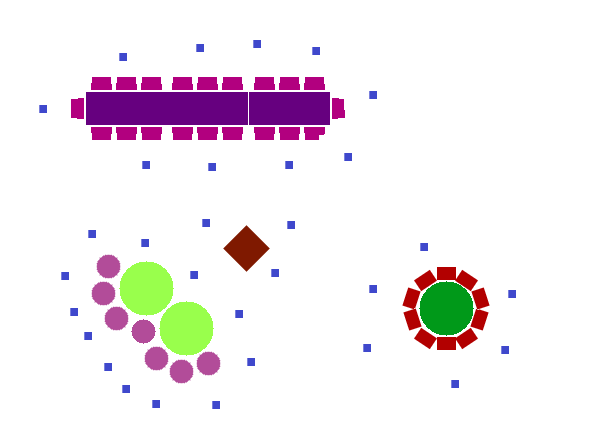}
	\caption{Data sampling points for training data acquisition. The blue points indicate the positions at which training data was sampled. The sampling points are close to the objects to gather a sufficient dataset.}
	\label{fig:gt_sample_map}
\end{figure}
\begin{figure}[!h]
	\centering
	\includegraphics[width=0.4\textwidth]{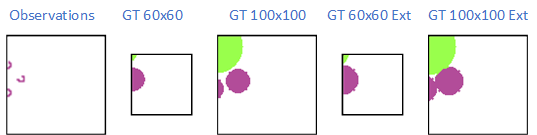}
 	\caption{Example of the dataset. The observation will be paired with four different groundtruths: the 60x60 normal ground truth,  the 100x100 normal ground truth, the 60x60 extended ground truth, the 100x100 extended ground truth.}
	\label{fig:gt_showcase}
\end{figure}

\noindent The training procedure will be taken offline to ensure efficiency. First, the training dataset will be created in the simulation environment. We have created three training maps within the simulation environment and collected training data from them. The training data contain 2 different categories, namely chair, and table. Totally 10 variants of the objects are included. The points from which training data is sampled are always in near proximity to the obstacles and shown in Fig. \ref{fig:gt_sample_map}. The semantic laser scan input $M_{scan}$, will be generated by the semantic labeling procedure described in section \ref{sec:data_acquisition}. For each $M_{scan}$, a paired ground truth for the local occupancy maps will be recorded. In this paper, we evaluated two different sizes for the local costmap, namely 60x60 and 100x100, while the robot is always in the center of the local map. For the training, we also have another set of data called extended ground truth. The object in the extended ground truth will be dilated by 5 pixels. An example is shown in Fig. \ref{fig:gt_showcase}

\noindent To mitigate the issue of an imbalanced dataset due to more free space than occupied space in the dataset, the loss function and the weight matrix $W_\alpha$ described in \cite{DBLP:journals/corr/abs-2104-03638} is used.
Once the dataset is processed, is it given as input to the model. 
As we said in section \ref{sec:data_acquisition}, our objective is to extend the imagination beyond the unseen area. 
The model is trained using different sizes of the input laser scan map

\begin{figure*}[!h]
	\centering
	\includegraphics[width=0.9\textwidth]{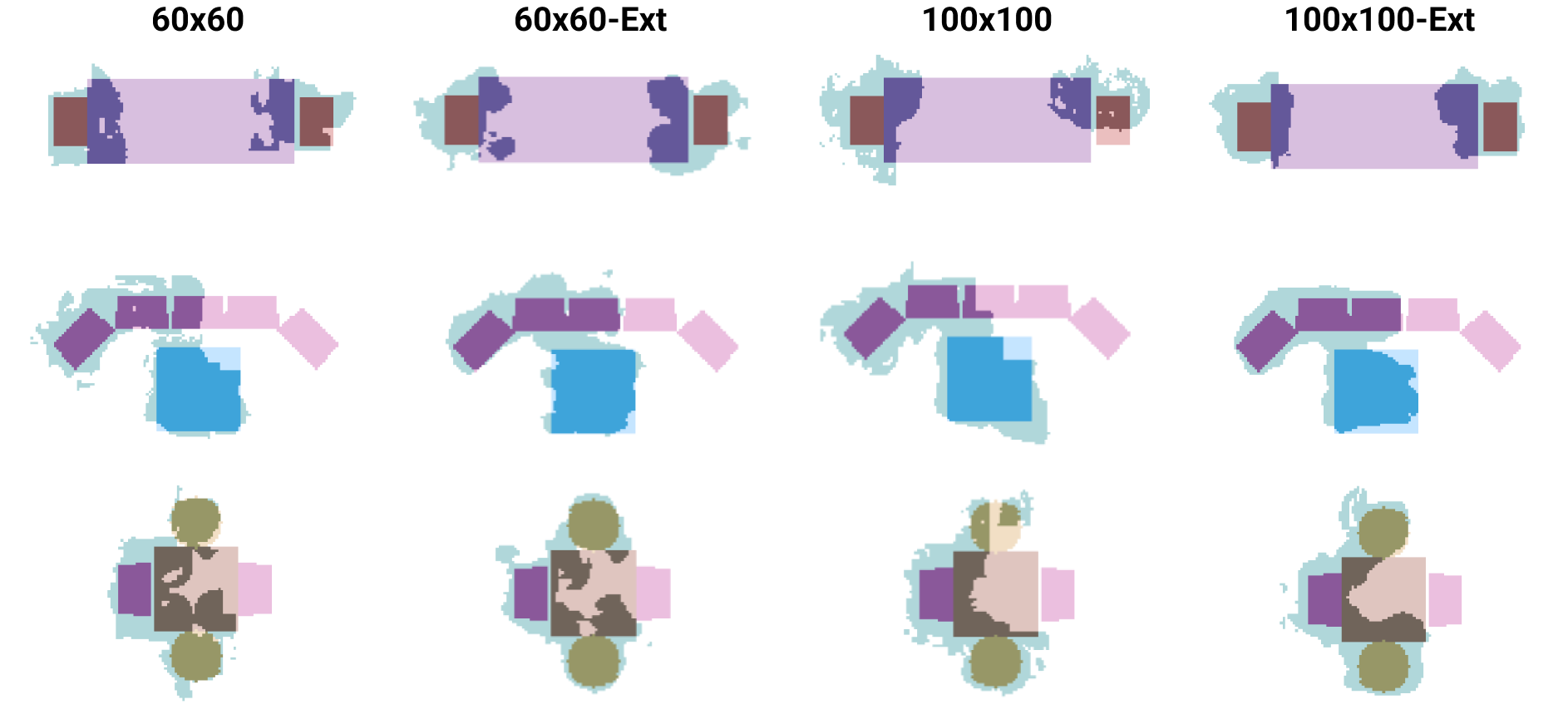}
	\caption{Imagination results for three different scenarios from four different models. The area covered by the blue mask is the imagination part. The results from four different models are trained on the different ground truth, namely 60x60, 60x60 Extended, 100x100, 100x100 Extended. The imaginations are based on thr observation and close to the real objects. The models trained with extended ground truth (denoted with 'Ext') predict the object more aggressively.}
	\label{fig:imagination}
\end{figure*}

\begin{table*}[!h]
\begin{center}
\caption{\label{tab:imagination} The number of imagination pixels from different imagination models }    

\begin{tabular}{ |c|c|c|c|c|c| } 

 \hline
 Imagination & 60x60 & 60x60 Ext & 100x100 & 100x100 Ext \\ 
 \hline
Imagination pixels & 32570 & 36529 & 32391 & 37754 \\ 
\hline
In object imagination pixels & 23616 & 24322 & 21737 & 26110\\ 
\hline
Out object imagination pixels & 8954 & 12207 & 10654 & 11644\\ 
 \hline
\end{tabular}
\end{center}
\end{table*}

\subsection{Integration into the ROS navigation stack}
\noindent To enhance observations and improve navigation, the imagination unit is integrated into the ROS navigation stack consisting of a global and local planner. Both planners work with ROS costmaps which are occupancy grid maps and represent areas that are not accessible by the robot. Especially the local costmap is dependent on the laser scan observations and is generated locally to react for dynamic changes and unknown areas not covered or considered by the global costmap. Both costmaps are consumed by a global and local planner respectively to plan and execute a path for the robot to avoid occupied areas. 
Hence, to improve these planners, the costmaps are enhanced with information gathered from our imagination unit. 
To validate our approach, we utilize a A* global planner \cite{hart1968formal} and a TEB local  planner \cite{rosmann2015timed}.

\begin{figure}[]
	\centering
	\includegraphics[width=0.4\textwidth]{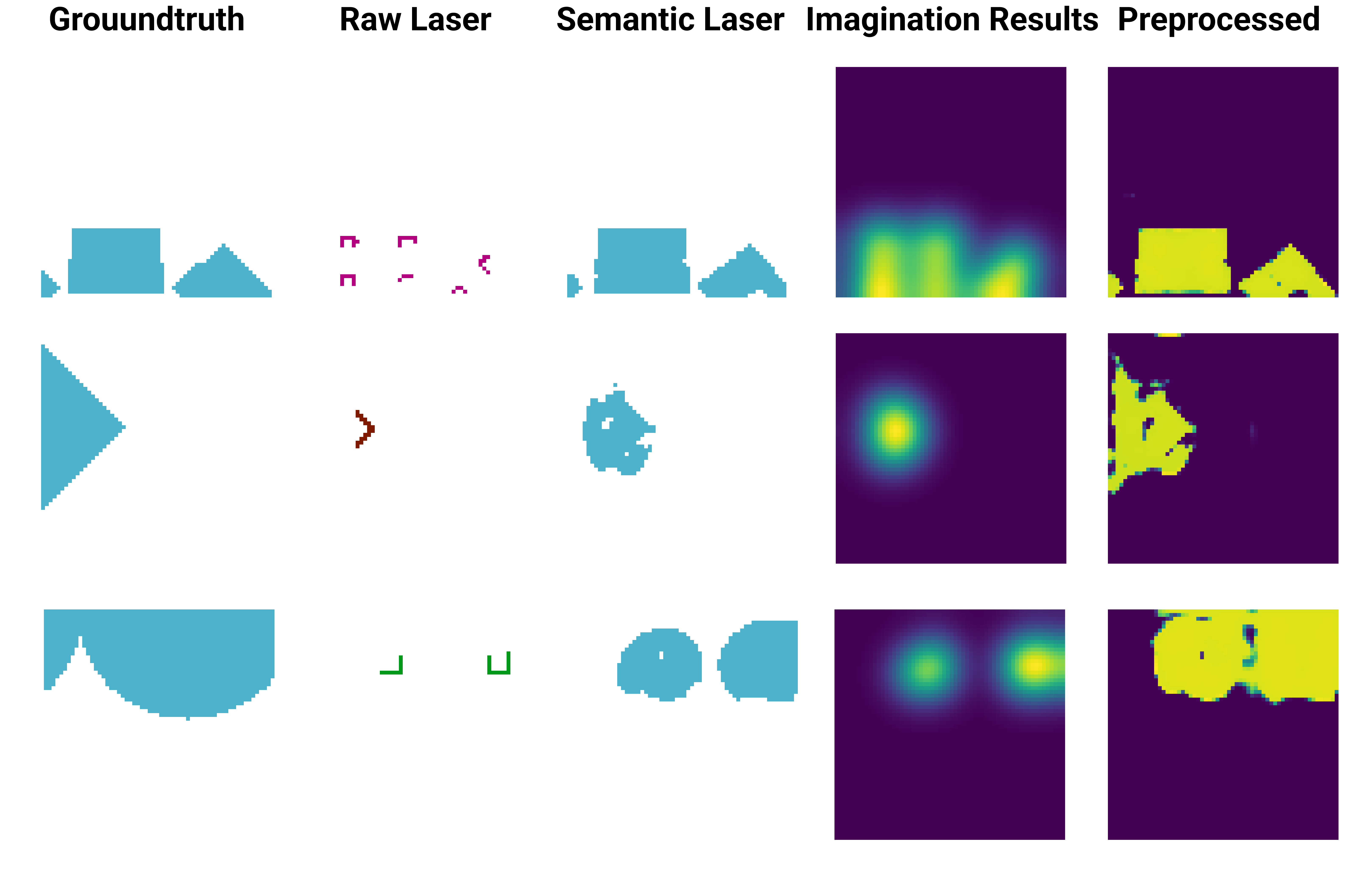}
	\caption{Imagination results together with ground truth image, laser scan, semantic laser scan, preprocessed mask, and imagination result for different example local areas.}
	\label{example_imagination_local_map}
\end{figure}

\begin{figure*}[!h]
	\centering
	\includegraphics[width=0.99\textwidth]{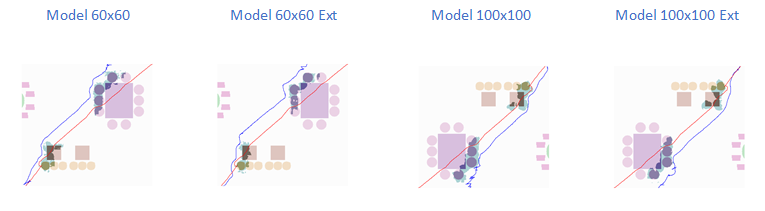}
	\caption{Trajectories of different agents. The red line indicates the path made by the agent without imagination while blue line indicates the paths by the agents with different imagination models. The squares and circles in different color represent different objects like chairs and tables. The models using extended observations are denoted with 'Ext'.}
	\label{fig:quality}
\end{figure*}

\begin{table*}[!h]
\caption{\label{tab:path_quantitative} Quantitative evaluation of the navigation}  
 
\begin{center}
\begin{tabular}{ |c|c|c|c|c|c|c| } 
\hline
 Metric & 60x60 & 60x60 Ext & 100x100 & 100x100 Ext & Without Imagination \\ 
 \hline
trajectory length in m & 36.15  & 29.29  & 32.64  & 27.97 & 21.5  \\ 
\hline
Duration(second) & 245.33  & 113.76  & 245.4  & 181.55 & 84.38 \\ 
\hline
Average velocity & 0.15 & 0.26 & 0.13 & 0.15 & 0.25\\ 
 \hline
\end{tabular}
\end{center}
\end{table*}

\section{Results and Evaluation}
\noindent We evaluated different agents with and without different imagination models to assess the effect of our different imagination models. A simulation test containing 16 paths in three different test scenarios is performed.

\subsection{Imagination Results}
\noindent Fig. \ref{example_imagination_local_map} illustrates the imagination masks, the raw imaginations (pre-processed), the imagination results (post-processed) together with the respective 2d laser scan observations, and the ground truth of three local areas of the test map. It is observed that the raw data created by the Unet decoder can already provide an accurate result. The mask is generated based on the raw laser data, which makes the prediction around the laser scan more reliable. The results demonstrate that the network can successfully predict the shape and the angle of the object based on the semantically labeled laser data.

\noindent Fig. \ref{fig:imagination} illustrates the imagination results of different scenarios and different models. We can observe that the imagination shows the correct position of the objects and enrich the sensor observation. To understand the difference of the imagination behavior between the different models, we calculate the total number of pixels of the imagination for three different evaluation scenarios and split them into two different classes, namely the in-object imagination pixels and the out-object imagination pixels. The results are listed in Table \ref{tab:imagination}. The model trained with extended ground truth predicts more pixels outside the object, while the model trained with 100x100-Ext ground truth data predicts significantly more pixels than other models. This is due to the fact that the model performs the imagination within a 100x100m area, although the observation is only limited in the size of 60x60m. The extended ground truth also encourages the model to classify the points close to the object also as occupied.

\subsection{Navigation Results}
\noindent To qualitatively assess the planners' behavior using the additional information of the imagination unit, we plotted the trajectories of the planners with and without imagination. Fig. \ref{fig:quality} depicts the paths created by the different agents. It is clearly observed that the agent without imagination module will go through the object for the shortest path. At the same time, all the agents with an imagination module choose the path without driving beneath the objects. The imagination can significantly influence the path chosen by the planner. Based on the local cost map and global cost map with the imagination, it has the ability to observe a wider occupancy from limited sensor observations, which can be essential for high-level semantic tasks such as avoiding sensitive or hazardous areas. Although the sensor data suggests that the path is safe to go, the imagination module will stop the agent from navigating through the object. Although the imagination areas from the different models have differences, the paths from these models have no significant changes. 

\noindent To compare the trajectories and kinematics of all models, we also perform a quantitative evaluation for the paths. Table \ref{tab:path_quantitative} lists the results. It shows that the length of the path has increased with the imagination module, since it will not go through the object. On the other hand, the average velocity of the planner is significantly lower. The possible reason is that the imagination interference with the path planner so it has to find a new sub-goal more frequently.

\section{Conclusion}

\noindent In this paper, we proposed an imagination method to enhance the robot's view based on 2D laser scan data. Therefore, we proposed data pre-and post-processing methods to semantically label 2d laser scan, map them into a semantic top-down map and filter out imagination noise. Subsequently, we trained an Unet-based neural network with different input sizes and demonstrated that the model could learn to detect and imagine objects based solely on 2D laser scan data. The resulting occupancy information is directly merged with the cost map of the ROS navigation stack to improve the planner's safety.
For evaluation, we integrated the imagination module with conventional planning algorithms and tested the systems on three different maps. Results show that the imagination module could influence the planner behavior by providing a wider field of view and avoiding the specified areas. This could be essential for robots with a limited observation space to maintain safe navigation or for high-level tasks such as avoiding pre-defined objects. Future works will focus on the integration of more obstacles for detection and imagination, as well as more extensive evaluations of different planners and maps. Moreover, we aspire to validate our approach on real robots.

\appendix
\noindent The code is publicly available at https://github.com/ignc-research/2d-imagination



\addtolength{\textheight}{-1cm} 




\typeout{}
\bibliographystyle{IEEEtran}
\bibliography{main}

\end{document}